%% file: arxiv_submit_manuscript.tex
\documentclass[journal]{IEEEtran}
\usepackage{graphicx}
\usepackage{subfigure}
\usepackage{subfloat}
\usepackage{booktabs}
\usepackage{picinpar}
\usepackage{float}
\usepackage{subfigure}
\usepackage{amsfonts,amssymb,amsbsy}
\usepackage{mathrsfs}
\usepackage{comment}
\usepackage{amsmath}
\usepackage{multirow}
\usepackage{cite}
\usepackage{algorithm}
\usepackage{algorithmic}

\begin{document}
\pdfoutput=1
%
\title{\huge \bf Indoor Image Representation by High-Level Semantic Features}

\author{Chiranjibi Sitaula, Yong Xiang \IEEEmembership{Senior Member, IEEE}, Yushu Zhang,
Xuequan Lu \IEEEmembership{Member, IEEE} and Sunil Aryal \IEEEmembership{Member, IEEE}
  
\thanks{C. Sitaula, Y. Xiang, and Y. Zhang, X. Lu, and S. Aryal are with School of Information Technology, Deakin University, Victoria 3125, Australia.
}     
}

\vspace{-0.35in}

\markboth{}
{Shell \MakeLowercase{\textit{et al.}}: Bare Demo of IEEEtran.cls for Journals}
%

\maketitle

\begin{abstract}
Indoor image features extraction is a fundamental problem in multiple fields such as image processing, pattern recognition, robotics  and so on. Nevertheless, most of the existing feature extraction methods, which extract features based on pixels, color, shape/object parts or objects on images, suffer from limited capabilities in describing semantic information (e.g., object association). These techniques therefore involve undesired classification performance. To tackle this issue, we propose the notion of high-level semantic features and design four steps to extract them. Specifically, we first construct the objects pattern dictionary through extracting raw objects in the images, and then retrieve and extract semantic objects from the objects pattern dictionary. We finally extract our high-level semantic features based on the calculated probability and delta parameter. Experiments on three publicly available datasets (MIT-67, Scene15 and NYU V1) show that our feature extraction approach outperforms state-of-the-art feature extraction methods for indoor image classification, given a lower dimension of our features than those methods.

\end{abstract}



\maketitle

\input{introduction.tex}

\input{relatedwork.tex}

\input{method.tex}

\input{results.tex}

\input{conclusion.tex}

\section{Acknowledgment}
We would like to thank Mr. Rakesh K. Bachchan for helping us extract NYU V1 dataset from the repository.

\bibliographystyle{IEEEtran}
\bibliography{reference}
\end{document}

%% file: introduction.tex
\section{Introduction}
\label{sec:introduction}
\IEEEPARstart{I}{mage} recognition and classification has remained an active research field. It has a wide range of applications \cite{foresti2005active} such as robotics, object recognition, object localisation, video surveillance, and so on. To perform the task of image recognition and classification, we usually need to represent each image
by a set of features. Generally, there are three categories of image features: low-level, middle-level, and high-level features.

Low-level features \cite{zeglazi_sift_2016, oliva2005gist,oliva_modeling_2001,dalal2005histograms,wu_centrist:_2011,xiao_mcentrist:_2014,song2014low,margolin2014otc,feng2018finding} are typically extracted using pixels, colour intensity or texture of the image. These features lack the spatial information on the image, thereby deteriorating the classification accuracy, especially for scene images (indoor/outdoor images). To improve the classification performance
of low-level features, middle-level features were proposed. Middle-level features \cite{lin_learning_2014, juneja2013blocks, parizi2012reconfigurable, lazebnik2006beyond} contain spatial information that yields features of certain parts or shapes on the image. Indoor images often involve one or multiple objects which can intuitively assist in recognising the categories of images. Thus, the object-level information for the objects on images could enhance the classification accuracy. 
 The middle-level features are limited in depicting objects on images while high-level features \cite{li2010object,tang_g-ms2f:_2017,cheng_scene_2018} including objects can do so.
 High-level features are considered as the prominent features for the images, including objects on indoor/outdoor images \cite{li2010object,tang_g-ms2f:_2017,cheng_scene_2018}. 
 In other words, high-level features can represent an image with the help of object details. Despite that high-level features are more powerful than middle-/low-level features, they still have limited performance for indoor images which often involve multiple objects with associations. 
\begin{figure}[b]
\begin{center}
\includegraphics[width=3cm, height=30mm,keepaspectratio]{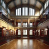}
\includegraphics[width=3cm, height=30mm,keepaspectratio]{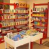}
\caption{Two library images that look dissimilar.}
\label{fig:1}
\end{center}
\end{figure}

Domain-specific features are the specific types of features that are designed at the specific domain. In our work, the features representing semantic objects and their associations are domain-specific features.
Domain-specific features are important to solve the specific classification problem. As an example, most of the low-level features, which are based on color or intensity as their features, have poor performance for indoor images involving multiple objects\cite{zeglazi_sift_2016}.

Indoor images are challenging because they usually include associated objects. For instance, two indoor categories library and kitchen may contain similar table and desk objects, but how can we differentiate the images? 
Similarly, we can hardly find the similarity using traditional features such as \cite{zeglazi_sift_2016}
if two images of the library category have different structures and the same types of objects (e.g., Fig. \ref{fig:1}). As a result, it is difficult to classify indoor images. 
In addition, recent existing methods such as \cite{8085139}, \cite{tang_g-ms2f:_2017} hardly solve the inter-class similarity and intra-class dissimilarity issues 
The recent work called objectness \cite{cheng_scene_2018} considered objects and their associations to some extent for indoor images. However, their research yielded high dimensional features which creates burden in image classification. 


Motivated by the above issues, we introduce the notion of semantic features that are calculated based on the semantic objects. General syntactic features may not solve these issues because the semantic meaning varies in different scenarios. The context, rather than objects, can define their separability in such cases. For example, the presence of books and tables in two different scenarios like kitchen and library would make machine ``confused'' in classification. While domain or context based objects (i.e., semantic) can help enhance classification in such cases. 
Semantic objects are the representative objects which are typically extracted from the object pattern dictionary of the corresponding category\cite{cheng_scene_2018}. 
These objects are retrieved by mapping raw object tags of the input images with the co-occurrence pattern of raw object tags in the corresponding object pattern dictionary. Note that raw object tags are the deep tags which are extracted from the ImageNet pre-trained  deep learning model. 
When retrieving the semantic objects for the raw objects through pattern dictionary, we can detect their co-occurrence patterns in the corresponding category with the help of rules we designed. Suppose we extract the semantic objects like books and chairs from the image, the co-occurrence of books and chairs will be higher in the library category than other categories. Similarly, the co-occurrence pattern of microwave and bread is higher in kitchen than other categories. The association information of objects can help extract meaningful features (i.e., semantic) on images. Also, the introduced semantic features can also alleviate inter-/intra-class (dis)similarity issues, because we can always extract meaningful information for whatever types of global layouts (similar or dissimilar).

To extract semantic features, we propose two main steps for our approach. We first design the object pattern dictionaries 
for each category, and then extract the semantic objects of each image according to their probabilities in the category. The rules are defined to exploit the co-occurrence patterns for the corresponding dictionaries. 
After the extraction of semantic objects with their probabilities in the corresponding category, we calculate semantic features by using the probability and delta parameters in various categories.
This is based on the motivation that the importance of objects differs in different categories. For example, the importance of book and desk is higher in library than other categories such as restaurant. 

The \textit{main contributions} of this paper are summarised as follows.
\begin{enumerate}
\item  We design the object pattern dictionary which encodes the  associations of indoor raw objects for each category. With the help of this dictionary, the semantic objects of the candidate objects for each image are calculated by mapping the candidate objects of the image with the dictionary. Four propositions are proposed to handle this procedure.
\item  We calculate the high-level semantic features with the help of semantic objects. The computed features usually have a low dimensional size. 
We perform a fusion of the probability and delta parameters to explore the prominent high-level semantic features. 
\item The introduced  high-level  semantic  features  are  tested  on three different publicly available datasets (MIT-67\cite{quattoni_recognizing_2009}, Scene15\cite{fei-fei_bayesian_2005}, and NYU V1\cite{silberman_indoor_2011}) for the task of classification. Experimental results show that our features are effective and outperform existing features in terms of classification accuracy. 
\end{enumerate}

%% file: relatedwork.tex
\section{Related Work}
Low-level features are extracted based on the pixels and color on the images. 
Some popular low-level features are Scale-Invariant Feature Transform (SIFT) \cite{zeglazi_sift_2016}, Generalized 
Search Trees (GIST) \cite{oliva2005gist,oliva_modeling_2001}, Histogram of Gradient (HOG) \cite{dalal2005histograms}, CENsus TRansform hISTogram (CENTRIST) \cite{wu_centrist:_2011}, multi-channel (mCENTRIST) \cite{xiao_mcentrist:_2014} and OTC \cite{margolin2014otc}. 
Since these features \cite{zeglazi_sift_2016, oliva2005gist, oliva_modeling_2001, dalal2005histograms, wu_centrist:_2011, xiao_mcentrist:_2014, margolin2014otc,feng2018finding} exploit the local information on the image, they provide neither the global structural information nor the object information. They may not work properly if the image feature extraction needs global structural details of the image. As a result, the classification accuracy of those features would also be low.
The classification accuracy can be improved, especially for indoor/outdoor images, if we can represent an image in different way such as edges, shapes, and parts\cite{katole2015hierarchical}.
This leads to middle-level features.

Middle-level features are extracted from the intermediate layers of the deep learning model. They can also be extracted by the traditional methods using parts or regions \cite{lin_learning_2014, juneja2013blocks, parizi2012reconfigurable, lazebnik2006beyond} on the image. Recent extraction methods of deep learning based middle-level features are bilinear \cite{7968351}, Deep Un-structured Convolutional Activations (DUCA) \cite{khan_discriminative_2016}, Bag of Spatial Parts (BoSP) \cite{8085139}, Locally Supervised Deep Hybrid Model (LS-DHM) \cite{guo_locally_2017} and so on. 
In the bilinear approach, the middle-level features from two deep learning models were fused with the help of the outer product to extract the final features of the image. These features were used for the classification. Similarly, in DUCA, the features of the $7$\textsuperscript{th} layer of the deep learning model were used as the middle-level features which showed that the features have higher discriminability in the classification than the low-level features. The BoSP model considered the features of pooling layers ($4$\textsuperscript{th} and $5$\textsuperscript{th} layers) as the middle-level features. Last but not the least, another method called LS-DHM was proposed, which exploited a hybrid model for the extraction of the middle-level features with the help of $4$\textsuperscript{th} layer of deep learning model in a 7-layers AlexNet \cite{krizhevsky2012imagenet}.
These features are extracted based on the lines, segments, shapes, and parts of the objects in the image. Thus, the classification accuracy using these features is higher than the low-level features because they are extracted in a higher level beyond the pixel and textural level. These features also provide certain semantic information of the objects in the image.
Different hierarchical layers provides different types of features in deep learning. We obtain more semantic information related to objects of the image while extracting features from the intermediate layers\cite{8085139,guo_locally_2017}.

Spatial units, extracted from their intermediate layers, are  fundamental in feature maps of deep learning models. For instance, if we have a feature map of 7*7*512 size extracted from the intermediate layers, the number of spatial units is 49, each with 512-D feature size.
Although the intermediate layers provide more semantic information with the help of their spatial units on different feature maps, 
these features can hardly obtain full semantic information of the objects in the image. This demands the use of features in a higher level (i.e., high-level features). 

We retrieve the high-level features from the top-layers (probability layers and $FC$-layers) of the deep learning model. Similarly, these features can also be extracted based on the traditional methods like Object Bank \cite{li2010object,song2014low} in which the features are extracted with the aid of object properties. 
Regarding indoor/outdoor images, objects-based features are very important because of the presence of objects and their associations in the image. It is very difficult to represent these associations by the help of low-level and middle-level features. Since high-level features are based on objects, we introduce prominent features related to the objects (i.e., high-level semantic features).
Recent high-level features are GMS2F \cite{tang_g-ms2f:_2017} and Objectness \cite{cheng_scene_2018}. 

%% file: method.tex
\section{Proposed Method}
The proposed method 
comprises the following steps,
namely: A) objects pattern dictionary construction, B) semantic objects extraction, C) probability and delta parameter calculation, and D) extraction of our high-level semantic features. For image classification, the high-level semantic features of the images are normalized in the attribute level before feeding them into the Support Vector Machine (SVM). Section III-A and III-B and III-C are the processing
steps for the extraction of semantic objects and Section D is the feature extraction step. 

\subsection{Objects Pattern Dictionary}
We design a domain-specific dictionary (i.e., objects pattern dictionary) which helps to explore the pattern of the objects occurring in the indoor images. Quite different from the Natural Language Processing (NLP) dictionary \cite{miller1995WordNet,kipper2006extending} and the sparse coding dictionary \cite{mairal2009online}, our dictionary demonstrates the relationship of indoor objects in the indoor scenes.
The objects pattern dictionary is to extract the semantically related objects on the image under the corresponding category of indoor scene images.  
To construct the objects pattern dictionary for each category, we select a fixed number of images per category and perform two steps: image slicing, and raw objects extraction and dictionary building.

\subsubsection{Image slicing}
\begin{figure}[t]
\center
 \includegraphics[width=0.45\textwidth]{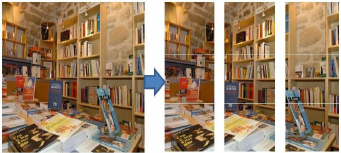} 
  \caption{Slicing of the image.}
 \label{fig:2}
\end{figure}

We slice each image into different sub-images to focus on the objects information. With the assistance of different sub-images, different objects and their frequencies are recorded. The occurrence of these frequencies of objects in the image help to reveal the semantic relationship of objects. Let ${{I_k}}$ be the ${k^{th}}$ image. Then, the sub-images of the $k^{th}$ image are represented by ${\{S^k_l\}_{l=1}^n}$, where $n$ is the number of sub-images per image.
The size of the dictionary depends on the number of sub-images per category. We construct three differing sizes of the dictionary to evaluate the robustness. For MIT-67 dataset, we randomly select 100 images per category and slice each into different numbers of sub-images, such as 9, 16, and 25. The number of sub-images per image determines the number of objects in the dictionary (i.e., dictionary size). The 9, 16 and 25 sub-images per image will yield the dictionary sizes of 9000, 16000 and 25000, respectively. We re-scale the original images into a suitable size before slicing. 
Fig.~\ref{fig:2} demonstrates the slicing of the an image into 9 sub-images to extract the raw objects. 
\begin{figure}[b]
\center
 \includegraphics[width=0.50\textwidth]{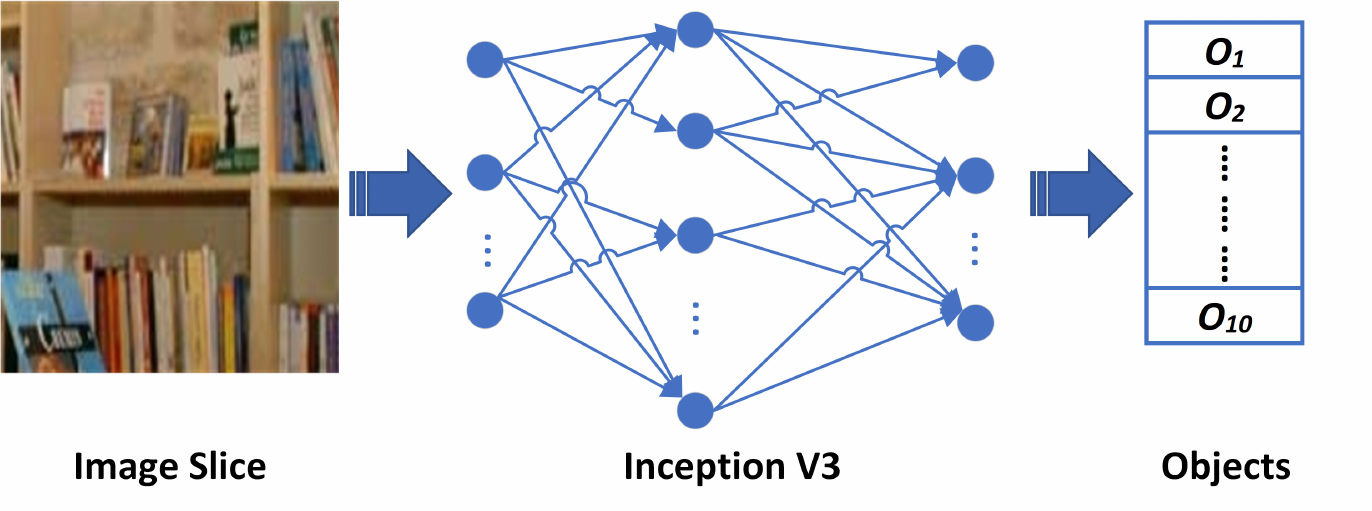}
  \caption{Block diagram for the extraction of raw objects of the image.}
 \label{fig:3}
\end{figure}

\subsubsection{Raw Objects Extraction and Dictionary building}
After slicing every image into sub-images, each sub-image will be re-scaled to feed the pre-trained deep learning model, Inception V3 \cite{szegedy_rethinking_2016} for the extraction of the names of the involved objects. 
This pre-trained model was trained with the ImageNet \cite{deng_imagenet:_2009} dataset that contains 1000 object categories. We choose Inception V3 because of three reasons: a) it has a deeper architecture than VGG-Net and AlexNet and has the capability to produce more semantic information on the image through its deeper layers, thereby helping to extract the more accurate names of objects. b) The computational complexity of this model is found to be lower than VGG-Net and other state-of-the-art deep architectures\cite{szegedy_rethinking_2016}.
c) The error rate of this model is lower than the state-of-the-art deep learning models such as GoogleNet, VGG-Net, and Inception-V2 \cite{szegedy_rethinking_2016}. The output of the multinomial distribution of deep learning model is shown in \eqref{eq:0}.
\begin{equation}
{P(y=c|z)}=\frac{e^{z_k}}{\sum_j e^{z_j}},
\label{eq:0}
\end{equation}
where $z$ is the multinomial probability which is extracted from the Softmax layer and $c$ is one among $1000$ categories for the ImageNet based pre-trained deep learning models.
Among the 1000 objects, 
we mainly consider the top ten raw objects with high probability scores in this work. The rest of the objects are ignored as they are less related to the images. Selecting the ``best'' objects from the image helps to exploit the distinguishable features.
Fig. \ref{fig:3} shows the extraction of the top ten raw objects of the 
sub-images based on the pre-trained Inception V3 model.
After the extraction of raw objects, our objective is to design the semantic dictionary, which is named objects pattern dictionary. 

To design this dictionary, we construct a raw dictionary first to list the raw objects of the indoor images using concatenation operation.
\begin{algorithm}[tb]
\caption{Raw Dictionary} 
\begin{algorithmic}[1]
\REQUIRE Set of objects for each category ${I^i_{k_{o_j}}}$
\ENSURE Raw dictionary for each category ${{D_i}}$
\STATE \textbf{for} $i=1$ to $m$
\STATE $\mspace{15mu}$  \textbf{for} $j = 1$ to $n$ 
\STATE $\mspace{30mu}$    \textbf{for} $k=1$ to $p$
\STATE  $\mspace{45mu} {{D_i}} \leftarrow \bigcup_{i,j,k}{I^i_{k_{o_j}}}$
\STATE $\mspace{30mu}$ \textbf{end for}
\STATE $\mspace{15mu}$ \textbf{end for}
\STATE \textbf{end for}
\end{algorithmic}
\label{algo:1}
\end{algorithm}
Algorithm \ref{algo:1} shows how to design a raw dictionary using the objects of the sub-images
extracted by the operation in Fig. \ref{fig:3}. We denote the number of categories by $m$\textsuperscript{th}, the number of images by $n$\textsuperscript{th}  and the number of objects by $p$\textsuperscript{th} 
. Similarly, ${I^i_{k_{o_j}}}$ represents the object $o_j$ for the ${i^{th}}$ category on the $k^{th}$ image,  and ${D_i}$ is the raw dictionary of $i^{th}$ category. ${I^i_{k_{o_j}}}$ is extracted using sub-images ${\{S^k_l\}_{l=1}^n}$ of the $k^{th}$ image and its output order for each category. Here, the raw dictionary is defined as the list of ordered objects collected from some indoor images of the corresponding category.
The raw dictionary is designed according to the objects' output order of the Inception V3 model.
The objects are concatenated exactly in the same order as obtained from Inception V3 model. This is helpful to show the relationship of the objects in the category.


While assembling those objects, the order shows certain associations among objects in the image. Furthermore, the list contains nine thousand objects in total if we consider one hundred images per category. The size of the dictionary is determined with the number of images available for different datasets. For instance, MIT-67 and Scene15 contain sufficient images to make dictionary using 100 images per category. It does not apply to NYU V1 dataset as some categories in NYU V1 dataset contain less than 100 images. 
\begin{algorithm}[t]
\caption{Objects Pattern Dictionary} \label{1}
\begin{algorithmic}[1]
\REQUIRE Raw dictionary ${{D_i}}$
\ENSURE Objects pattern dictionary ${{C_i}}$
\STATE \textbf{for} $i=1$ to $m$
\STATE $\mspace{15mu}$  \textbf{for} $k = 1$ to $n$ 
\STATE  $\mspace{30mu} $Occurrence of objects i.e., [{$o_k$},
{$o_{k+1}$}] and [{$o_{k-1}$}, {$o_{k}$}]
\STATE $\mspace{15mu}$ \textbf{end for}
\STATE \textbf{end for}
\STATE \textbf{for} $i=1$ to $m$
\STATE $\mspace{15mu}$\textbf{for} $j=i$ to $n$
\STATE $\mspace{30mu} {{C_i}} \leftarrow  $Sort Descending order.
\STATE $\mspace{15mu}$ \textbf{end for}
\STATE \textbf{end for}
\end{algorithmic}
\label{algo:2}
\end{algorithm}
After the construction of the raw dictionary, we refine it further to design the objects pattern dictionary. The objects pattern dictionary is based on the semantic relatedness of the objects. To explore the semantic relatedness of the objects, we investigate the object co-occurrence pattern property for those images.

Object co-occurrence is the main component of the objects pattern dictionary. To detect the co-occurrence of the objects in the image, we study the adjacent object pairs and their co-occurrence
in the image of the category. 
In order to solve the co-occurrence problem, we design Algorithm \ref{algo:2} to extract
the frequency (i.e., the degree of relationship) of the adjacent pairs of objects. Here, $D_i$ and $C_i$ represent the raw dictionary and the objects pattern dictionary for the $i^{th}$ category, respectively. For each category, the objects pattern dictionary is designed based on the objects and orders of the raw dictionary. The objects are selected using forward and backward directions. Inspired by the 2-gram model \cite{brown1992class} in Natural Language Processing, we utilize the adjacent pair of objects which co-occur in the image. If they have high-frequency pairs in the categories, higher
degree of relationship exists between the objects. Unlike the 2-gram model that considers only the previous gram, we take the previous and next gram of the corresponding object as the semantic objects to design the objects pattern dictionary. For example, if we take an object $o_i$, then its relationship can be shown with lower indexed object $o_{i-1}$ and higher indexed object $o_{i+1}$.
The general structure of the objects pattern dictionary (${C_i}$) which shows the semantically related objects for the particular category is shown in \eqref{eq:2}. 
\begin{equation}
{C_i}=
  \begin{pmatrix}
  \{{o_1},{o_2}\}\rightarrow {n_1}\\
  \{{o_2},{o_3}\}\rightarrow {n_2}\\
  \{{o_4},{o_5}\}\rightarrow {n_3}\\
  .\\
  .\\
  .\\
  \{{o_m},{o_n}\}\rightarrow {n_k}\\
  \end{pmatrix},
  \label{eq:2}
\end{equation}
where ${n_1}$, ${n_2}$, ${n_3}$...${n_k}$ indicate the frequency of adjacent pairs obtained from the raw dictionary, $D_i$. For example, ${n_1}=C\{o_1, o_2\}$ where $C$ is the count of object pair in the raw dictionary ($D_i$). 
The semantic relationship of objects and their occurrence are stored in the key-value pair format. We will apply our proposed propositions (Section \ref{sec:semanticobjectextraction}) on the objects pattern dictionary, which contains the order pattern of objects, to extract the semantic objects for the image. 
This domain-specific dictionary is to extract the semantically related objects for those types of images.

\subsection{Semantic Objects Extraction}
\label{sec:semanticobjectextraction}
After the design of objects pattern dictionary for each category, the semantic objects extraction step needs to be conducted for each image under the corresponding category. In this step, we retrieve the semantic objects from the objects pattern dictionary of the corresponding candidate objects in the image. To extract the semantic objects of the image, we slice every image into different sub-images (such as 9, 16, and 25) to extract the raw objects. The highly frequent raw objects of the image from multiple sub-images are selected as the candidate objects to map the corresponding objects pattern dictionary. 
 we propose four propositions to facilitate the extraction of the semantic objects. Each proposition is stated and proved.
\subsubsection{Proposition 1}
If $o_1$ and $o_2$ are co-occurring in the multiple sub-images ${{I_j}}$, then they can be used interchangeably.

\textbf{Proof.}
If two objects are not co-occurring in the sub-images, we can say that these objects are unrelated to each other. For instance, if two objects $o_1$ and $o_2$ appear together, we say that one's presence is related to others presence. If these objects are not co-occurring in the sub-image, we say that they are mutually exclusive (the presence of one object is not related to other objects presence).
This claims that the co-occurring objects unveils their associations in the image.

\subsubsection{Proposition 2}
If two pairs of objects ($o_1$, $o_2$) and ($o_1$, $o_3$) are co-occurring in the multiple sub-images ${{I_j}}$, there exists the relationship between $o_2$ and $o_3$.

\textbf{Proof.}
If two objects $o_1$ and $o_2$ are co-occurring in the sub-images frequently, we can claim that they are correlated each other. Similarly, if the objects $o_1$ and $o_3$ are also co-occurring in the sub-images, then the associations between them can be claimed. Furthermore, from these two associations, we see that $o_1$ is correlated with both $o_2$ and $o_3$. Hence, we can prove that $o_2$ and $o_3$ are related to each other.

\subsubsection{Proposition 3}
If ($o_1$, $o_2$) is co-occurring in the image sub-images and ($o_1$, $o_3$) and ($o_2$, $o_4$) are also occurring, it can be proved that $o_3$ and $o_4$ are related.

\textbf{Proof.}
If two objects are co-occurring in the sub-images, the occurrence shows the relationship clearly between them which can be proved from the Proposition 1 that the usage of one in place of another makes no difference. In the above objects pair, the first pair shows that $o_1$ is related to $o_2$ and they can be used interchangeably.  Similarly, in the second pair ($o_1$, $o_3$), $o_1$ and $o_3$ are related. Furthermore, the pair ($o_2$, $o_4$) also shows that $o_2$ and $o_4$ are also related. In this way, it can be proved that $o_2$ and $o_3$ are related to each other from Proposition 2.
\subsubsection{Proposition 4}
If ($o_1$, $o_2$) and ($o_2$, $o_3$) are co-occurring in the image, it shows the relationship between $o_1$ and $o_3$.

\textbf{Proof.}
If ($o_1$, $o_2$) show that the two objects are related to each other, they can be used interchangeably. It shows the relationship between these two objects. Similarly, the pair ($o_2$, $o_3$) shows that there exists a relationship between these objects in the image. Looking in both pairs, there will be the occurrences of objects ($o_1$, $o_2$, and $o_3$) in the sub-images, which proves that $o_1$ and $o_3$ are related to each other.
\begin{algorithm}[t]
\caption{Semantic Objects Extraction} \label{1}
\begin{algorithmic}[1]
\REQUIRE Objects pattern dictionary ${{C_i}}$. \\
              $\mspace{20mu}$ Candidate objects of the image ${{I_{k_{o_j}}^{i}}}$ 
\ENSURE Semantic objects ${{W_{k_{o_j}}^{C_i}}}$
\STATE  \textbf{for} $i = 1$ to $n$ 
\STATE  $\mspace{15mu}$\textbf{for} $j = 1$ to $m$ 
\STATE $\mspace{30mu}$ ${{W_{k_{o_j}}^{C_i}}} \leftarrow{{I_{k_{o_j}}^{i}}}{\Phi} {{C_i}} $
\STATE $\mspace{15mu}$ \textbf{end for} 
\STATE \textbf{end for}
\end{algorithmic}
\label{algo:3}
\end{algorithm}

In Algorithm \ref{algo:3}, ${{\Phi}}$ represents the mapping function of the selected raw objects with the objects pattern dictionary to produce semantic objects of the image. Mathematically, it is written as ${\Phi}:{I}\times{C}\to {W}$. This function applies the four propositions listed above. The ${{\Phi}}$ function can be any proposition we proposed.
In the ${{\Phi}}$ function, $I$ represents the notation for the image, $C$ represents the objects pattern dictionary, and $W$ represents the extracted semantic objects.
 Similarly, ${{W_{k_{o_j}}^{C_i}}}$ is the list of semantic objects for the objects ${o_j}$ of ${I_k}$ under ${C_i}$ object pattern dictionary.
 
For explanation purposes, let  the image ${{I_k}}$ contain two candidate objects in ${{I_{k_{o_j}}^{i}}}$ such as $o_1$ and $o_2$ which uses objects pattern dictionary $C_i$. We search the related objects of $o_1$ in $C_i$ with the aid of the proposed propositions, and extract the co-occurring pairs. We select highly frequent co-occurred objects that are related to the candidate objects of the image under the corresponding category. For instance, let us consider a dictionary from Eq. \eqref{eq:2} as an objects pattern dictionary.
 \begin{align*}
&{{W}}\{{o_1}\}=\{{o_2}, {o_3}\}(\textit{Proposition~1, and~Proposition~4}),\\
&{{W}}\{{o_2}\}=\{{o_3}\}(\textit{Proposition~1}), \\
&\textnormal{finally}, {{W}}\{{o_1}, {o_2}\}=\{{o_3}\},
\end{align*}
where we only select the unique objects that does not belong to the raw objects of the image.
The extracted semantic objects are stored in ${{W_{k_{o_j}}^{C_i}}}$. Here, ${o_3}$ is the semantic object. We use a number of
candidate objects to extract the corresponding semantic objects in the image.

\subsection{Probability and Delta Parameter Calculation}
\label{sec:delparameters}
After the extraction of semantic objects of the image, the probabilities and delta parameters of the semantic objects need to be calculated. The raw dictionary is used to calculate the probabilities. 
Denote each object by $o_j$ and the raw dictionary by $D$\textsubscript{\textit{i}}.
\begin{equation}
V=\bigcup_{i,j}\Delta_{i_j}*p({o_j}|{D_i})\\
\label{eq:4}
\end{equation}
\begin{equation}
\Delta_{i_j}=\frac{f_{o_j}^{D_i}}{c({o_k}^{D_i})}
\label{eq:5}
\end{equation}
${p}(o_j|D_i)$ is the probability of the object ($o_j$) in a different dictionary ($D_i$). Similarly,
$f_{o_j}^{D_i}$, $c({o_k}^{D_i})$ and  $\Delta_{i_j}$ are the frequency of an object ($o_j$), the total number of objects and the delta parameter value of the $j^{th}$ object, in the dictionary ($D_i$)
The fusion of the probability and delta parameter is performed via Eq.~\eqref{eq:4}. 
The delta parameter is the primary factor that helps distinguish the images having inter-class similarities. We design six different types of delta parameters in our experiments.
\begin{figure*}[t]
\center
 \includegraphics[width=0.90\textwidth, keepaspectratio]{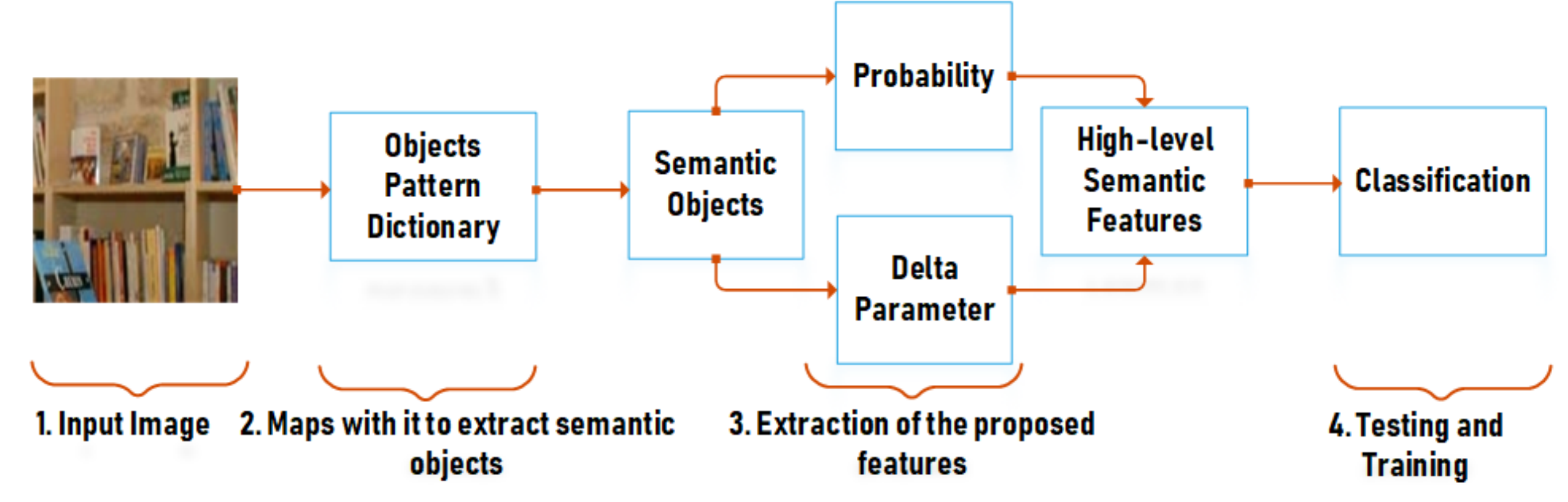}
 \caption{Detailed diagram for the extraction of high-level semantic features and classification by Sequential Minimal Optimization (SMO) based Support Vector Machine. }
  \label{fig:6}
  \end{figure*}

\subsubsection{Normal Delta Parameter}
This is the normal delta parameter defined in Eq.~\eqref{eq:5}. This is a normal probability function of the objects belonging to the category.

\subsubsection{Avg Delta Parameter}
The average delta parameter measures the impact of normal probability with respect to the total probabilities of all semantic objects. It is shown in the Eq. \eqref{eq:6}.
\begin{equation}
\Delta_{i_j}=\frac{p({o_j}|{D_i})}{\sum_{i,j}{p({o_j}|{D_i})}}
\label{eq:6}
\end{equation}

\subsubsection{Normalized Delta Parameter}
Eq. \eqref{eq:7} is the normalized delta parameter. To make probability non-zero and divide by zero exception handling, we add $1$ to all the frequency count operation if the dividing by zero exception occurs.
\begin{equation}
\Delta_{i_j}=\frac{p({o_j}|{D_i})}{\sqrt[4]{p({o_j}|{D_i})}}
\label{eq:7}
\end{equation}

\subsubsection{Multi-probability Delta Parameter}
This delta parameter is the result of multiplying the normal probability value with the frequency of the objects in the corresponding category. The multi-probability delta parameter is defined in Eq. \eqref{eq:8}, where $f({o_j})$ represents the frequency of an object, ${o_j}$.
\begin{equation}
\Delta_{i_j}=p({o_j}|{D_i})*f({o_j})
\label{eq:8}
\end{equation}

\subsubsection{Root-based Delta Parameter}
This type of delta parameter is obtained by taking the square root of the normal probability. The main objective of this delta parameter is to test the efficacy of increased normal probabilities.
The square root of probability between 0 and 1 gives higher values. 
The root-based delta parameter is shown by the Eq. \eqref{eq:9}.
\begin{equation}
\Delta_{i_j}=\sqrt{p({o_j}|{D_i})}
\label{eq:9}
\end{equation}

\subsubsection{Decimal Scaling or Divide Delta Parameter}
In this parameter, the probability score is made smaller than the original value. The number of decimal values increases with the help of this parameter. This type of delta parameter is to study the effect of lower probability scores. To perform the decimal scaling delta parameter, we use Eq. \eqref{eq:10}.
\begin{equation}
\Delta_{i_j}=\frac{p({o_j}|{D_i})}{10^k}, k={0,1....n}
\label{eq:10}
\end{equation}

\begin{figure}[b]
\centering
\includegraphics[width=0.45\textwidth]{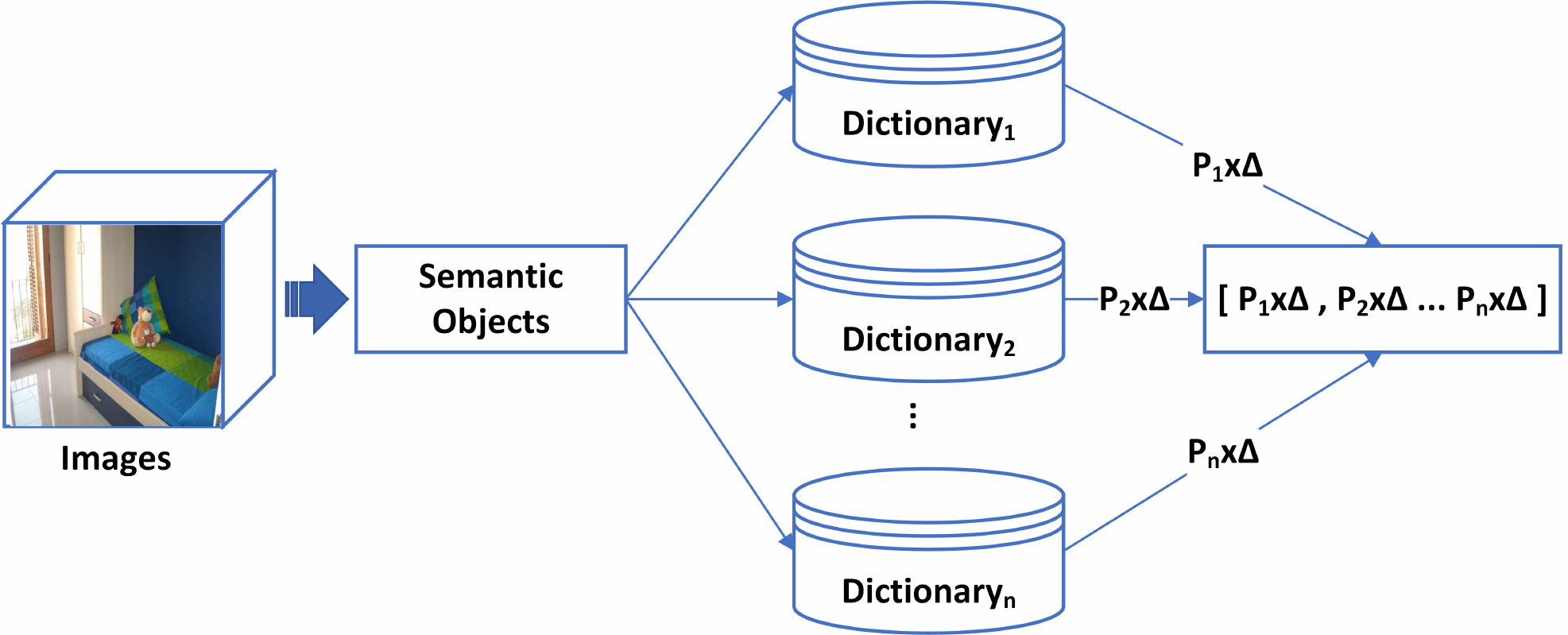}
\caption{Block diagram of the proposed model for the extraction of high-level semantic features. 
The dictionary in the diagram represents the raw dictionary for each category.}
 \label{fig:5}
\end{figure}
\subsection{Extraction of High-level Semantic Features}
The high-level semantic features are extracted by the help of Eq. \eqref{eq:4}, which is the product of the probability and delta parameter. Six different delta parameters (Eq. \eqref{eq:5} - Eq. \eqref{eq:10}) are tested one by one to choose the best delta parameter. Among these six delta parameters, we found that the normal delta parameter is better when designing the high-level semantic features. The resulting features have a higher classification accuracy of indoor images involving inter-/intra-class structural dissimilarities.
The extraction of semantic objects and high-level semantic features 
extraction flow can be seen in Fig. \ref{fig:5}. 

\section{Analysis of  Proposed Method}
We use the pre-trained deep learning model to extract the objects only as the first step of the research.
Our method takes more time complexity in objects pattern dictionary construction and semantic objects extraction step. We study the time complexity of each operation while executing the proposed method.
The time complexity of the raw dictionary module is approximately quadratic  for $m$ categories, since it needs $p$ objects to form the dictionary from $n$ images of each category. As a result, the complexity is $O(nmp)$. 
In the objects pattern dictionary module, we need to find the adjacent objects occuring together. We need to track the adjacent objects and sort them. The worst complexity for dictionary construction is $O(n(m-1))$, and for sorting, the complexity does not exceed $O(nm^2)$. 

The extraction of semantic objects is the most expensive, where we need to search the objects pattern dictionary in the backward or forward direction, depending on the situation. If the proposition is simple like Proposition 1, the complexity is $O(nk(m-1))$. Here, $k$ is the number of candidate objects of the image for the extraction of semantic objects. We set a lower value of $k$, so that the complexity does not go higher. 
For the worst case, we should find the semantic objects moving in both directions using Proposition 2, 3 or 4 and the complexity is $O(nkm^2)$. Similarly, for the delta parameters and probability calculation, the complexity is $O(sn)$, where $s$ represents the extracted semantic objects. 
It is better to analyze the performance in terms of the worst cases because sometimes we need to perform expensive search operations of semantic objects. The overall time complexity of our approach is not greater than $O(nmp)$+$O(n(m-1))$+$O(nm^2)$+$O(nkm^2)$+$O(sn)$. Here, $m$ is higher than other values, $s$ is a constant equal to 5 and $k$ is also a constant value which can be adjusted by users. 
Now the worst complexity of the algorithm becomes $O(m^3)$. The size of objects pattern dictionary $m$ is the main driving factor of the complexity.
For training and testing using the SVM, the complexity is determined by the number of categories and dimension of the feature instances. 

%% file: results.tex
\section{Experiments and Analysis}

\subsection{Datasets}
We conduct experiments on three popular datasets: MIT-67 \cite{quattoni_recognizing_2009}, Scene15 \cite{fei-fei_bayesian_2005} and NYU V1 \cite{silberman_indoor_2011} datasets. 
Among these datasets, the MIT-67 and NYU V1 are indoor scenes datasets, whereas the Scene15 is a combination of indoor and outdoor scenes. As suggested in \cite{jin_rank_2013}, we use the same number of images for the NYU V1 dataset. We extract the proposed features and performed classification using the given train/test split ratio of each dataset. 

\subsubsection{MIT-67}
MIT-67 includes $15,620$ images for $67$ classes (categories) in total. 
The MIT-67, the biggest dataset used in this work, has been used in many previous methods such as ROI with GIST \cite{quattoni_recognizing_2009}, MM-Scene \cite{zhu_large_2010}, 
 Object Bank \cite{li2010object}, RBoW \cite{parizi2012reconfigurable}, BOP \cite{juneja2013blocks}, 
OTC \cite{margolin2014otc}, ISPR \cite{lin_learning_2014}, CNN-MOP \cite{gong_multi-scale_2014}, DUCA \cite{khan_discriminative_2016}, BoSP \cite{8085139}, Bilinear \cite{7968351}, G-MS2F \cite{tang_g-ms2f:_2017}, VSAD \cite{wang_weakly_2017}, Objectness \cite{cheng_scene_2018} and so on.
In the experiment, we design 10 sets of train/test dataset. For this, we select 100 images randomly from each category 
and split them into the $8:2$ ratio (train/test ratio) to use in the experiment.
We repeated this technique 10 times to design 10 sets train/test data for the experiment.


\subsubsection{Scene15}
This dataset includes $15$ categories, where some categories are outdoor. It contains $4,485$ images in total. The Scene15 dataset has been used in methods like GIST-color \cite{oliva_modeling_2001}, 
 SPM \cite{lazebnik2006beyond}, CENTRIST \cite{wu_centrist:_2011}, OTC \cite{margolin2014otc}, ISPR \cite{lin_learning_2014}, G-MS2F \cite{tang_g-ms2f:_2017}, DUCA \cite{khan_discriminative_2016}, Objectness \cite{cheng_scene_2018} and so on. 
While selecting the training and testing images for the research, we choose 100 images randomly for training and remaining images for testing, which is a standard protocol for this dataset. We randomly design 10 sample sets for this dataset in this way for the research.

\subsubsection{NYU V1}
It comprises of $2,284$ images and $7$ indoor categories. 
The NYU V1 dataset was used in works such as BoW with SIFT \cite{silberman_indoor_2011}, RGB with LLC \cite{jin_rank_2013}, RGB-LLC-RPSL \cite{jin_rank_2013} and DUCA \cite{khan_discriminative_2016}.
For the experiments, we take $6:4$ as the train/test split ratio for each category. We randomly design $10$ sample sets for this dataset as well. 

\subsection{Implementation Details}
Firstly, the sub-images are generated by the image slicer library of the python programming language \cite{pythoncite}.
Each input image should be in 3-channel (RGB) format and feed into the pre-trained deep learning models. The images from two datasets (MIT-67 and NYU V1) are already in 3-channel format, while the images of Scene15 dataset are in grayscale format. We use keras \cite{chollet2017kerasR} to convert grayscale images into the 3-channel format. Their algorithm repeats three times to get a 3-channel image for a grayscale image. The objects are then extracted by the Inception V3 model implemented on the popular keras \cite{chollet2017kerasR} library in R \cite{rcite}. These objects are processed to get semantic objects. The proposed feature extraction operations based on semantic objects are implemented using Python.
\begin{table}[tb] \quad
\caption{Comparison of our proposed features with features from the state-of-the-art approaches on MIT-67.}
\label{tab:1}
\centering
\begin{tabular}{p{5cm} p{2.6cm}}
\toprule
Method & Accuracy(\%)\\
\midrule
ROI with GIST\cite{quattoni_recognizing_2009} &26.1 \\
MM-Scene\cite{zhu_large_2010} & 28.3\\
Object Bank\cite{li2010object}&37.6\\
RBoW\cite{parizi2012reconfigurable}&37.9\\
BOP\cite{juneja2013blocks}&46.1\\
OTC\cite{margolin2014otc}&47.3\\
ISPR\cite{lin_learning_2014}&50.1\\
CNN-MOP\cite{gong_multi-scale_2014} & 68.0\\
DUCA\cite{khan_discriminative_2016} & 71.8\\
BoSP\cite{8085139} & 78.2\\
Bilinear\cite{7968351} & 79.0\\
G-MS2F\cite{tang_g-ms2f:_2017}& 79.6\\
VSAD\cite{wang_weakly_2017}& 86.2\\
Objectness\cite{cheng_scene_2018}& 86.7\\
\textbf{Ours}& \textbf{94.1}\\
\hline
\end{tabular}
\end{table} 

To evaluate the proposed features for classification, the SVM based on SMO \cite{platt1998sequential} is used under Weka \cite{hall_weka_2009}. 
We employ a $10$-fold cross-validation approach for training with the default parameter setting of the SVM algorithm available in Weka. The detailed flow is illustrated in Fig. \ref{fig:6}.
We also do the ablation study for three key elements: dictionary size, number of slices and delta parameters. 

\subsection{Comparison With State-of-the-Art Features}
The quantitative comparisons of the proposed features with previous features are listed in Tables \ref{tab:1}, \ref{tab:2} and \ref{tab:3}. 
Tables \ref{tab:1}, \ref{tab:2} and \ref{tab:3} represents the performance on the MIT-67, Scene15 and NYU V1, respectively.
For fair comparisons, we utilize the same dataset and use the reported performance results for the previous approaches.
To estimate the classification accuracy on each dataset, we design 10 samples, each of which has a train/test split.
The average accuracy of samples on each dataset is used to compare with the state-of-the-art features. 
Compared with those existing features, we obtain noticeably higher classification accuracies on all datasets used in the research. 

While observing in Table \ref{tab:1}, we see that our proposed features yield a substantially higher classification accuracy on MIT-67. At the very beginning of the research on this dataset, the GIST \cite{quattoni_recognizing_2009} approach with the traditional low-level feature representation by ROI gives only $26.1\%$. Object Bank \cite{li2010object},
RBoW\cite{parizi2012reconfigurable}, BOP \cite{juneja2013blocks}, OTC \cite{margolin2014otc} and ISPR \cite{lin_learning_2014} provided accuracies of $37.6\%$, $37.9\%$, $46.1\%$, $47.3\%$, and $50.1\%$, respectively. The features based on traditional computer vision methods do not show promising results. 
These research works, focused on the handcrafted technology, have larger feature dimensions for the representation of the image. Their features simply rely on the low-level components such as colors or pixels of the image which may be not suitable for the images.
The classification accuracy surged higher after adopting CNN-based techniques. The CNN-MOP\cite{gong_multi-scale_2014} approach obtains an accuracy of $68\%$, which is over twice of the accuracy produced by ROI with GIST. 
With the help of middle-level features based on deep features, the accuracy is improved drastically in classification because of the representation using parts of the objects in the image. 
The features extracted by DUCA \cite{khan_discriminative_2016} approach outperforms the normal CNN-MOP approach, which follows the proper step-wise operations of the feature extraction. This approach yields an accuracy of $71.8\%$. 
\begin{table}[t] 
\caption{Comparison of our proposed features with the features from state-of-the-art approaches on Scene15.}
\label{tab:2}
\centering
\begin{tabular}{p{5cm} p{2.5cm}}
\toprule
Method & Accuracy(\%)\\
\midrule
GIST-color\cite{oliva_modeling_2001} &69.5 \\
SPM\cite{lazebnik2006beyond}&81.4\\
CENTRIST\cite{wu_centrist:_2011}&83.9\\
OTC\cite{margolin2014otc}&84.4\\
ISPR\cite{lin_learning_2014} & 85.1\\
G-MS2F\cite{tang_g-ms2f:_2017} & 92.9\\
DUCA\cite{khan_discriminative_2016} & 94.5\\
Objectness\cite{cheng_scene_2018}& 95.8\\
\textbf{Ours} & \textbf{98.9}\\
\hline
\end{tabular}
\end{table}
\begin{table}[t] 
\caption{Comparison of our proposed features with features from the state-of-the-art approaches on NYU V1.}
\label{tab:3}
\centering
\begin{tabular}{p{5cm} p{2.5cm}}
\toprule
Method & Accuracy(\%)\\
\midrule
BoW with SIFT\cite{silberman_indoor_2011} &55.2 \\
RGB with LLC\cite{jin_rank_2013} & 78.1\\
RGB-LLC-RPSL\cite{jin_rank_2013} & 79.5\\
DUCA\cite{khan_discriminative_2016} & 80.6\\
\textbf{Ours}& \textbf{96.5}\\
\hline
\end{tabular}
\end{table}
BoSP \cite{8085139}, which again considered middle-level features with spatial pooling layers, produces an accuracy of $78.21\%$. These features have a lower dimensional size than the previous features. The middle-level features from Bilinear \cite{7968351}, high-level features G-MS2F \cite{tang_g-ms2f:_2017} and Objectness \cite{cheng_scene_2018} give $79\%$, $79.63\%$, and $86.76\%$, respectively. This shows the effectiveness of high-level features on the MIT-67 dataset based on deep learning models. 
However, these features still suffer from a high dimensional cost for the image representation. By contrast, our high-level semantic features enable a lower feature size while a significantly improved accuracy ($94.1\%$).

Similarly, we see the promising accuracy of our proposed features on the Scene15 dataset. The accuracies of all features are listed in Table \ref{tab:2}. The low accuracy is obtained by the GIST-based low-level features which have a higher dimension. The features extracted by the traditional methods such as SPM \cite{lazebnik2006beyond}, CENTRIST \cite{wu_centrist:_2011} and OTC \cite{margolin2014otc} yield accuracies of $81.4\%$, $83.9\%$ and $84.4\%$, respectively. Furthermore, middle-level features extracted by ISPR \cite{lin_learning_2014} show a promising result in terms of a classification accuracy of $85.1\%$ due to the part based representation of objects in the image. The accuracy is improved significantly with deep learning based features which involve hierarchical features of the image. The middle-level features from DUCA \cite{khan_discriminative_2016} approach provides an accuracy of $94.5\%$. The high-level features extracted by G-MS2F \cite{tang_g-ms2f:_2017} generates an accuracy of $92.90\%$. Our proposed features yield an average accuracy of $\textbf{98.9}\%$ which is the highest among the state-of-the-art features. 
\begin{figure}[b]
\centering
 \includegraphics[width=0.45\textwidth]{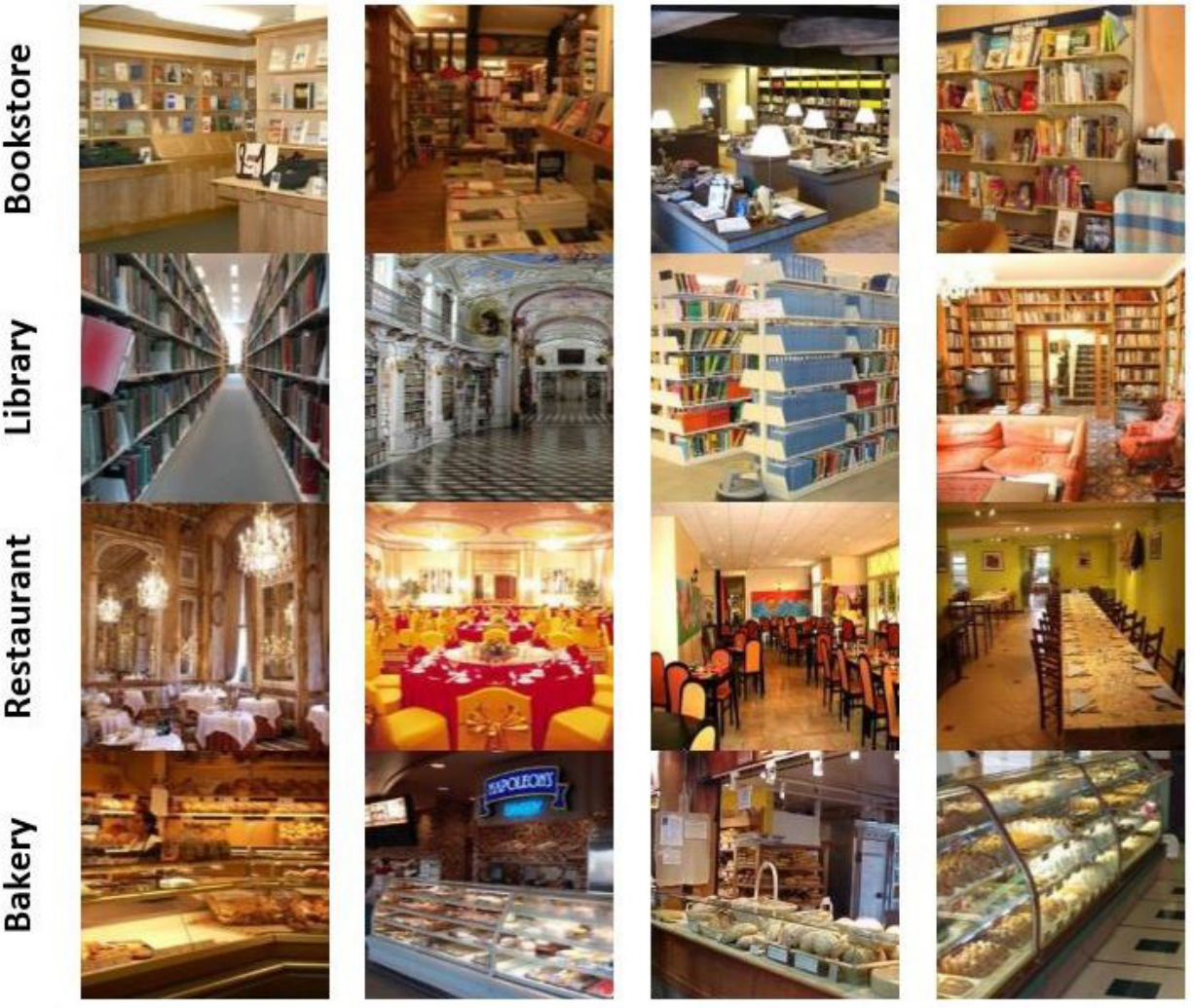}
   \caption{Sample images from MIT-67 dataset\cite{quattoni_recognizing_2009}.}
  \label{fig:7}
  \end{figure}
\begin{figure}[t]
\centering
 \includegraphics[width=0.43\textwidth, height=4cm]{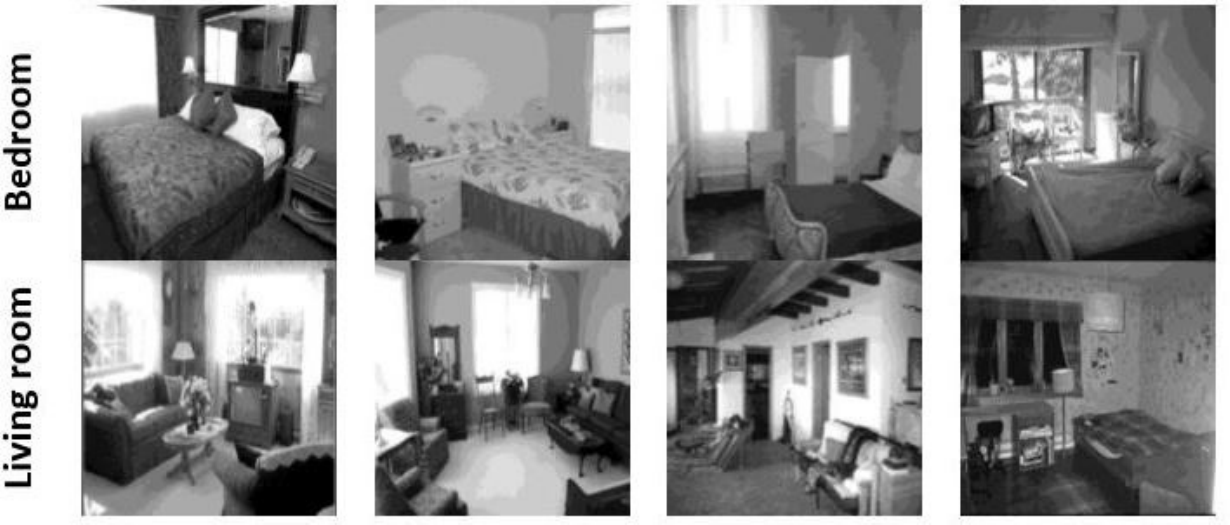}
  \caption{Sample images from Scene15 dataset\cite{fei-fei_bayesian_2005}.}
  \label{fig:8}
  \end{figure}
\begin{figure}[t]
\center
 \includegraphics[width=0.43\textwidth,height=4cm]{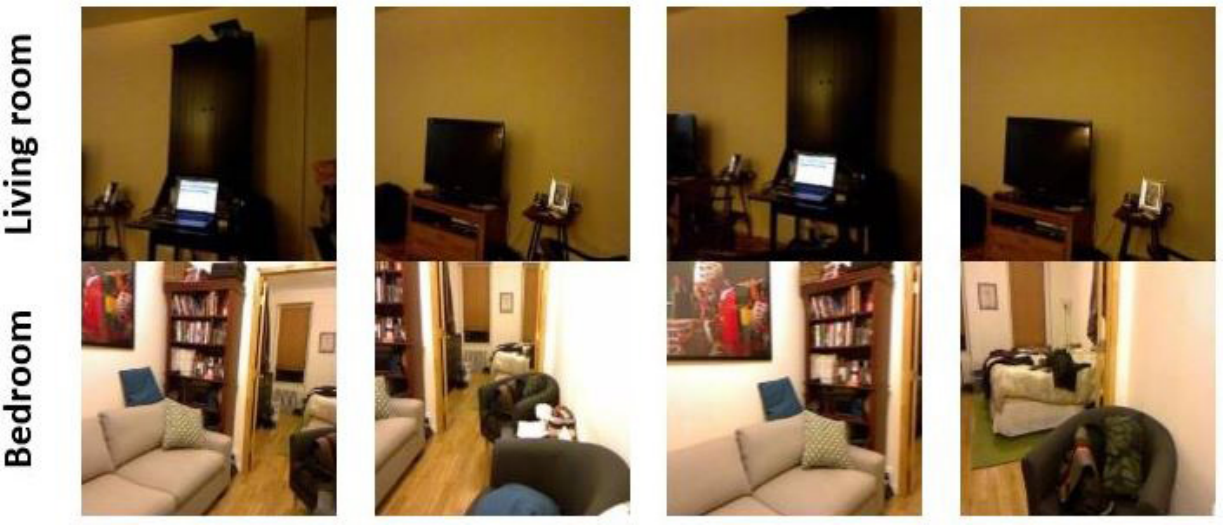}
   \caption{Sample images from NYU V1 dataset\cite{silberman_indoor_2011}.}
 \label{fig:9}
 \end{figure}

Table \ref{tab:3} also shows that deep learning based features can produce promising results in image classification for the NYU V1 dataset.  We noticed that the accuracy increases with the quality of the features set designed. The Bag of Visual Words (BoVW) approach with SIFT \cite{silberman_indoor_2011} features has a $55.2\%$ accuracy. The features based on deep learning yields a higher quality. Similarly, our proposed features outperforms all these existing features by achieving an average accuracy of $96.5\%$. 

Three datasets involve various images which have impact in designing features, thereby affecting the classification performance. Some images from different categories are similar in nature, as shown in Fig. \ref{fig:7} for MIT-67 dataset. The cross-class 
similarity deteriorates the classification performance. Fig. \ref{fig:7} shows the images of four categories i.e., bookstore, library, restaurant and bakery. We see the structural similarity of bookstore and library images. However, the structural dissimilarity of the images within the category can be also seen. 
Furthermore, the Scene15 dataset (Fig. \ref{fig:8}) has some challenging complex images having intra- and cross-class structural barriers. Similar to the MIT-67 dataset, this dataset also contains many 
images having such complexities.  
\begin{table}[tb] 
\caption{Classification accuracy (\%) of 10 different samples of each dataset.}
\label{tab:4}
\centering
\begin{tabular}{p{1.7cm} p{1.7cm} p{1.6cm} p{1.6cm}}
\toprule
Sample & MIT-67(\%) & Scene15(\%) & NYU V1(\%)\\
\midrule
1 &93.9 & 99.1 & 97.2 \\
2 & 93.5 & 99.2 & 94.9\\
3 & 92.0 & 98.7 & 97.3\\
4 & 95.1 & 98.9 & 95.3\\
5 & 94.7 & 99.1 & 96.9 \\
6 & 94.0 & 99.1 & 97.2\\
7 & 94.7 & 99.1 & 96.6\\
8 & 94.7 & 98.8 & 96.0\\
9 & 94.1 & 99.1 & 97.3 \\
10 & 95.1 & 98.6 & 96.3\\
\hline
\end{tabular}
\end{table} 
We also see the intra-class structural dissimilarity of the images in Fig. \ref{fig:9} for the NYU V1 dataset.
However, the images of this dataset contain fewer obstacles for the categories, compared to other datasets. Whatever types of datasets used in the experiments, each dataset has its own obstacles for the calculation of the proposed features. To generalize and ensure the quality of the proposed features under these obstacles, an intuitive way is to average the classification accuracy of more samples on each dataset. We conduct a more in-depth experiment on each sample of each dataset, to further evaluate the classification accuracy.
The accuracies of each sample for different datasets are listed in Table \ref{tab:4}. 
It shows the stability of the proposed features for each dataset. 
The average accuracies for 10 samples of MIT-67, Scene15 and NYU V1 dataset are $94.1\%$, $98.9\%$ and $96.5\%$, respectively.

\subsection{{Ablative Analysis of Dictionary Size}}
 The number of objects in the dictionary determines the size of the dictionary. We design three different sizes of dictionaries to evaluate the separability of the proposed features on the MIT-67 dataset. Three different sizes of dictionaries are $9,000$, $16,000$, and $25,000$. These dictionaries are used to construct the objects pattern dictionaries. We extract semantic objects based on those object pattern dictionaries of the corresponding category and calculate the proposed features using those objects. Here, we use the corresponding numbers of sub-images for each dictionary in extracting the proposed features. For instance, on the $9,000$-size, $16,000$-size, $25,000$-size dictionary, we use $9$, $16$, and $25$ sub-images per image, respectively. We design $10$ sample sets for the evaluation of the dictionary size. 
 Table \ref{tab:5} enlists the classification accuracy of the proposed features under different dictionary sizes in the experiment. 
 \begin{table}[t] 
 \caption{Classification accuracy (\%) of 10 different samples of MIT-67 dataset for three different size of dictionaries.}
\label{tab:5}
\centering
\begin{tabular}{p{1.7cm} p{1.7cm}p{1.7cm}p{1.7cm}}
\toprule
&\multicolumn{3}{c}{Dictionary size}\\
 Sample&\cline{1-3}
 &9000& 16000& 25000\\
 \midrule
1 &93.6 & 94.1 & 93.3 \\
2 & 94.7 & 94.0 & 93.8\\
3 & 94.0 & 93.7 & 93.6\\
4 & 94.3 & 94.2 & 93.5\\
5 & 93.2 & 93.3 & 92.8 \\
6 & 92.9 & 93.2 & 91.4\\
7 & 92.4 & 94.1 & 93.5\\
8 & 92.8 & 94.1 & 93.1\\
9 & 92.9 & 93.6 & 93.4 \\
10 & 93.5 & 92.3 & 92.0\\
\hline
Average&93.4&93.6 &93.0\\
\hline
\end{tabular}
\end{table}
While performing the individual dictionary size evaluation with the corresponding number of sub-images (3x3 sub-images for 9000-size dictionary, 4x4 sub-images for 16000-size dictionary, and 5x5 sub-images for 25000-size dictionary),
we noticed that the $16000$-size dictionary obtains the best accuracy result in the classification.
\begin{table}[htbp] 
\caption{Classification accuracy (\%) of different samples of MIT-67 dataset for three different number of sub-images with three different size of dictionaries.}
\label{tab:6}
\centering
\begin{tabular}{p{1cm} p{1.7cm} p{1.2cm}p{1.2cm}p{1.2cm}}
\toprule
&&\multicolumn{3}{c}{Dictionary size}\\
 Sample&Sub-images&\cline{1-3}
  &&9000& 16000& 25000\\
  \midrule
  &\multirow{1}{*}{9 (3*3)} &93.2 & 91.8 & 91.3\\
1&\multirow{1}{*}{16 (4*4)} &93.9 & 94.0 & 94.7\\
&\multirow{1}{*}{25 (5*5)}& 93.2 & 92.2 & 92.3 \\
\hline
&\multirow{1}{*}{9 (3*3)} &93.5 & 91.9 & 91.4\\
2&\multirow{1}{*}{16 (4*4)} &93.5 & 93.3 & 93.2\\
&\multirow{1}{*}{25 (5*5)}& 94.4 & 94.2 & 94.1 \\
\hline
&\multirow{1}{*}{9 (3*3)} &93.0 & 92.3 & 90.8\\
3&\multirow{1}{*}{16 (4*4)} &92.0 & 91.7 & 92.0\\
&\multirow{1}{*}{25 (5*5)}& 92.7 & 93.0 & 93.2 \\
\hline
&\multirow{1}{*}{9 (3*3)} &92.9 & 92.2 & 91.5\\
4&\multirow{1}{*}{16 (4*4)} &95.1 & 92.8 & 93.3\\
&\multirow{1}{*}{25 (5*5)}& 92.7 & 92.3 & 92.2 \\
\hline
&\multirow{1}{*}{9 (3*3)} &93.2 & 92.0 & 91.7\\
5&\multirow{1}{*}{16 (4*4)} &94.7 & 93.8 & 93.2\\
&\multirow{1}{*}{25 (5*5)}& 93.8 & 93.2 & 93.5 \\
\hline
 &\multirow{1}{*}{9 (3*3)} &92.8 & 90.5 & 90.8\\
 6&\multirow{1}{*}{16 (4*4)} &94.0 & 94.7 & 94.0\\
&\multirow{1}{*}{25 (5*5)}& 94.2 & 93.5 & 93.5 \\
\hline
&\multirow{1}{*}{9 (3*3)} &93.8 & 91.9 & 90.7\\
7&\multirow{1}{*}{16 (4*4)} &94.7 & 94.7 & 94.4\\
&\multirow{1}{*}{25 (5*5)}& 95.0 & 94.1 & 93.8 \\
\hline
&\multirow{1}{*}{9 (3*3)} &93.6 & 92.4 & 92.5\\
8&\multirow{1}{*}{16 (4*4)} &94.7 & 94.3 & 93.5\\
&\multirow{1}{*}{25 (5*5)}& 93.8 & 93.3 & 93.8 \\
\hline
&\multirow{1}{*}{9 (3*3)} &93.8 & 92.4 & 91.4\\
9&\multirow{1}{*}{16 (4*4)} &94.1 & 93.0 & 93.3\\
&\multirow{1}{*}{25 (5*5)}& 94.3 & 94.0 & 92.9 \\
\hline
&\multirow{1}{*}{9 (3*3)} &93.5 & 91.5 & 91.2\\
10&\multirow{1}{*}{16 (4*4)} &95.1 & 94.7 & 94.1\\
&\multirow{1}{*}{25 (5*5)}& 93.1 & 92.2 & 92.2 \\
\hline
Average& &93.7 & 92.9 & 92.6\\
\hline
\end{tabular}
\end{table}

 \subsection{Ablative Analysis of the Number of Sub-images and Dictionary size}
 
 To analyze the effectiveness of the number of sub-images, 
 we exploit the relationship between the number of sub-images and the dictionary size for the proposed features. The numbers of sub-images per image used are 9, 16 and 25, respectively. Firstly, the semantic objects of each image are extracted using the corresponding dictionary. For example, for the images with 9 sub-images, we extract semantic objects using the $9,000$-size dictionary. The extracted semantic objects of each image are then utilized to respectively calculate the proposed features under three different dictionaries. Also, we design 10 sets of train/test data. The effectiveness of the number of sub-images is demonstrated in Table \ref{tab:6}. 
 The experiment reveals that the $9,000$-size dictionary is suitable for the proposed features extraction of these images. All three sub-images ($9$, $16$ and $25$) per image perform well on this $9,000$-size dictionary for extracting the features for the classification. This finding between the number of sub-images and the dictionary size helps to explore highly separable features for such type of images during the feature extraction.

\subsection{{Ablative Analysis of Delta Parameters}}

\begin{figure*}[t]
\center
 \includegraphics[width=\textwidth, height=5.3cm, keepaspectratio]{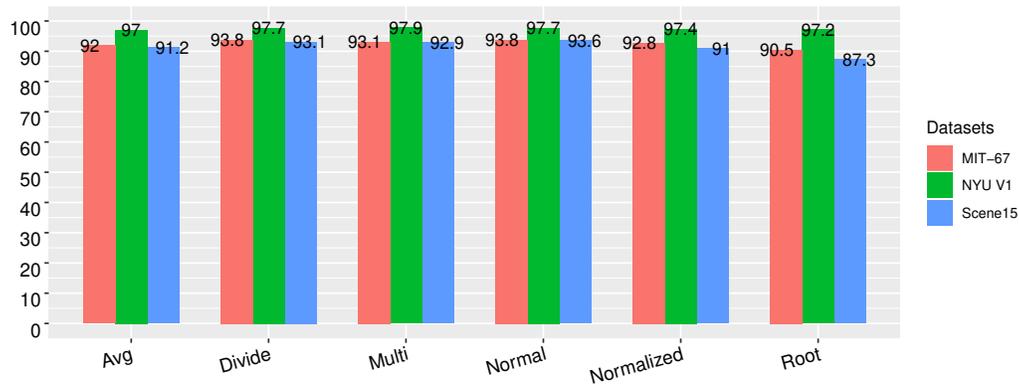}
  \caption{Comparison of six delta parameters on three datasets in terms of classification accuracy.}
 \label{fig:10}
 \end{figure*}

We designed six different types of delta parameters as the multiplier factors with the probability scores of the semantic objects in different categories. Since this parameter plays a crucial role in the design of the proposed features, we experiment them one by one on 
three datasets. We utilize the dictionary obtained from $3*3$ sub-images per image (e.g., $9,000$ for MIT-67 dataset) and corresponding semantic objects to evaluate the parameters. The details about these parameters are elaborated in Section \ref{sec:delparameters}. We use $9$ sub-images per image to analyze the delta parameters. 
We design the features based on each delta parameter on three datasets.
To test the robustness of the delta parameter, we design one set of data for each dataset by following the corresponding training and testing ratios. The accuracies are represented by the bar graph (Fig. \ref{fig:10}).  

While observing the individual classification accuracy in Fig. \ref{fig:10}, the accuracy of the normal delta parameter is higher than other delta parameters on MIT-67, Scene15 and NYU V1. However, the accuracy of 
divide delta parameter was the same as the normal delta parameter in terms of classification accuracy on the MIT-67.
The root-based delta parameter becomes worst for MIT-67 and Scene15 dataset. This result shows the robustness of the normal delta parameter.
 
Furthermore, we consider the average classification accuracy of each delta parameter on three datasets. The average accuracy of the proposed features that use a normal delta parameter on all three datasets is $95\%$, which is the highest accuracy. The lowest accuracy reported is the root-based delta parameter which achieves only $91.6\%$. 
 
 \begin{table}[t] 
\caption{Average classification accuracy (\%) of six delta parameters on MIT-67, Scene15, and NYU V1.}
\label{tab:7}
\centering
\begin{tabular}{p{1.3cm} p{0.6cm} p{0.6cm} p{0.6cm}p{0.7cm} p{1cm} p{0.7cm}}
\toprule
 {Delta parameters}& {Avg}& Divide& Multi&Normal&Normalized & Root\\
\midrule
Accuracy&{93.4}&94.8&94.6&\textbf{95}&93.7&91.6\\
\hline
\end{tabular}
\end{table}

%% file: conclusion.tex
\section{Conclusion}
We have proposed the high-level semantic features concept and designed a set of steps to extract them for the representation of the indoor images. The proposed features outperform the state-of-the-art features, in terms of indoor image classification. Our features have a lower dimension and higher separability than the existing features, thereby achieving higher classification accuracies. It has demonstrated that the semantic objects are important clues for extracting the image features with high separability. We believe this work will arouse new insights in the future. 